\pdfoutput=1
\documentclass[sigconf]{acmart}

\usepackage{colortbl}
\usepackage{balance}
\usepackage{multirow}

\usepackage{xcolor}

\AtBeginDocument{%
  }

\copyrightyear{2025}
\acmYear{2025}
\setcopyright{acmlicensed}
\acmConference[WWW Companion '25]{Companion Proceedings of the ACM Web Conference 2025}{April 28-May 2, 2025}{Sydney, NSW, Australia}
\acmBooktitle{Companion Proceedings of the ACM Web Conference 2025 (WWW Companion '25), April 28-May 2, 2025, Sydney, NSW, Australia}
\acmDOI{10.1145/3701716.3715557}
\acmISBN{979-8-4007-1331-6/25/04}

\begin{document}

\title{To Share or Not to Share: Investigating Weight Sharing in Variational Graph Autoencoders}


\author{Guillaume Salha-Galvan}
\affiliation{%
  \institution{Kibo Ryoku Research}
  \city{}
  \country{}
}
\email{gsalhagalvan@kiboryoku.com}

\author{Jiaying Xu}
\affiliation{%
  \institution{Kibo Ryoku Research}
  \city{}
  \country{}
}
\email{jxu@kiboryoku.com}

\renewcommand{\shortauthors}{Guillaume Salha-Galvan and Jiaying Xu}

\begin{abstract}
This paper investigates the understudied practice of weight sharing (WS) in variational graph autoencoders (VGAE). WS presents both benefits and drawbacks for VGAE model design and node embedding learning, leaving its overall relevance unclear and the question of whether it should be adopted unresolved. We rigorously analyze its implications and, through extensive experiments on a wide range of graphs and VGAE variants, demonstrate that the benefits of WS consistently outweigh its drawbacks. Based on our findings, we recommend WS as an effective approach to optimize, regularize, and simplify VGAE models without significant performance loss.
\end{abstract}



\maketitle

\section{Introduction}

Variational graph autoencoders (VGAE) have emerged as powerful unsupervised methods for graph representation learning~\cite{kipf2016-2,hamilton2020graph}. Combining an encoder-decoder architecture with a probabilistic framework to learn node embedding vectors, they have achieved success in downstream tasks central to Web applications, including link prediction and community detection~\cite{kipf2016-2,salha2022contributions,salhagalvan2022modularity,pan2018arga,grover2019graphite,choong2018learning,salha2019-2}.

VGAE models employ two encoders, typically graph neural networks~\cite{hamilton2020graph,kipf2016-1}, to learn the mean and variance of embedding vectors. An existing yet underexamined practice in these models is weight sharing (WS), which involves using the same weights in the hidden layers of both encoders~\cite{kipf2016-2}. As explained in Section~\ref{s2}, WS was introduced by default--and without strong justification--in popular VGAE implementations~\cite{kipf2016-2,pytorchgeometric}. Consequently, many works building on these implementations have also adopted WS, often without explicit acknowledgment.
Nonetheless, we argue that WS has significant implications for model design and embedding learning. While some may be beneficial, others could hinder model performance. As a result, the relevance of WS in VGAE models remains unclear, leaving the question of whether it should be adopted~unresolved.

In this paper, we address this question by conducting, to our knowledge, the first in-depth investigation of WS in VGAE models. We believe our study will benefit researchers considering WS for VGAE simplification but worried about its implications--particularly in Web applications, where both efficiency and performance are critical. Specifically, our contributions in this paper are as follows:
\begin{itemize}
\item We present a detailed analysis of why and how WS is used in VGAE, discussing its positive and negative implications.
\item We evaluate the relevance of WS through comprehensive experiments on 10 VGAE variants and 16 datasets of varying natures, sizes, and characteristics. Our results show that the advantages of WS consistently outweigh its drawbacks, with no significant performance loss observed in any experiment.
\item We publicly release our source code on GitHub to ensure reproducibility and support further exploration of WS in VGAE models: \href{https://github.com/kiboryoku/ws_vgae}{https://github.com/kiboryoku/ws\_vgae}.
\end{itemize} 

This paper is organized as follows. In Section~\ref{s2}, we introduce VGAE models and discuss the WS practice and its implications. We analyse our experiments in Section~\ref{s3}, and conclude in~Section~\ref{s4}.

\section{Investigating Weight Sharing in VGAE}
\label{s2}

We begin this section with preliminary background on VGAE and WS, before discussing the benefits and drawbacks of this practice.

\subsection{Background on VGAE}
\label{s21}

\subsubsection{Notation} We consider a graph $\mathcal{G} = (\mathcal{V}, \mathcal{E})$ with $n \in \mathbb{N}^*$ nodes and $m \in \mathbb{N}^*$ edges, and denote its $n \times n$ adjacency matrix (binary or weighted) as $A$. Each node $i \in \mathcal{V}$ may have a feature vector $x_i \in \mathbb{R}^f$, with $f \in \mathbb{N}^*$, providing information about it. These vectors form an $n \times f$ feature matrix $X$. For featureless graphs, we simply set $X = I_n$, the $n \times n$ identity matrix. 
Lastly, we use the term node embedding method to describe a technique representing each node $i \in \mathcal{V}$ with an ``embedding vector'' $z_i \in \mathbb{R}^d$ in a common vector space of dimension $d \in \mathbb{N}^*$, with $d \ll n$. Typically, these vectors aim to capture and summarize structural graph information~\cite{hamilton2020graph}.

\subsubsection{Node Embedding with VGAE}
Among the most popular unsupervised node embedding methods are variational graph autoencoders (VGAE), introduced by Kipf and Welling~\cite{kipf2016-2} as an extension of variational autoencoders (VAE)~\cite{kingma2013vae} to graphs. Unlike most node embedding methods, VGAE models learn an entire probability distribution $q(z_i)$ for each node, instead of a single vector $z_i$. This probabilistic approach captures uncertainty in the embedding space, providing greater flexibility in representing graph structures~\cite{salha2022contributions}.

VGAE-based embedding vectors have shown promising performance, enabling tasks such as link prediction and community detection~\cite{kipf2016-2,kipf2020phd, salha2020simple,li2020dirichlet,pan2018arga,grover2019graphite,choong2018learning,salha2019-2}. Due to space limitations in this short paper, we refer readers to the doctoral thesis of Salha-Galvan~\cite{salha2022contributions} for a detailed review of VGAE models and their applications.
While various model formulations are possible, the remainder of this Section~\ref{s2} focuses on the original VGAE by Kipf and Welling~\cite{kipf2016-2} for clarity, with alternative architectures being discussed in~Section~\ref{s3}.

\subsubsection{VGAE Encoding}
Kipf and Welling~\cite{kipf2016-2} sample each node embedding vector $z_i~\in~\mathbb{R}^d$ from a $d$-dimensional Gaussian distribution, characterized by a mean vector $\mu_i \in \mathbb{R}^d$ and a diagonal variance matrix $\Sigma_i = \text{diag}(\sigma_i^2)  \in (\mathbb{R}^+)^{d \times d}$, with $\sigma^2_i \in (\mathbb{R}^+)^d$. To learn these parameters, they employ two encoders. Letting $\mu$ and $\Sigma$ denote the $n \times d$ matrices containing the mean vectors and log-variance vectors for all nodes, respectively, they set:
\begin{equation}
\mu = \text{Encoder}_{\mu}(A,X) \text{ and } \Sigma = \text{Encoder}_{\Sigma}(A,X).
\label{neweq1}
\end{equation}

Specifically, Kipf and Welling~\cite{kipf2016-2} use graph convolutional networks (GCN) \cite{kipf2016-1} as encoders in their work. An $L$-layer GCN (with $L \geq 2$) consists of an input layer $H^{(0)}=X$, some hidden layer(s)
$H^{(l)} = \text{ReLU} (\tilde{A} H^{(l-1)} W^{(l-1)})$ for $ l \in \{1,...,L-1\}$, and an output layer $
H^{(L)} = \tilde{A} H^{(L-1)} W^{(L-1)}$.  Here,  
$\tilde{A} = (D+I_n)^{-\frac{1}{2}}(A+I_n)(D+I_n)^{-\frac{1}{2}}$ denotes the symmetric normalization of the adjacency matrix $A$, where $D=\mathrm{diag}(A \mathbf{1}_n)$ is the degree matrix of $A$ and $\mathbf{1}_n$ is the $n$-dimensional vector of ones~\cite{kipf2016-1}. At each layer, the GCN learns a vector for each node by aggregating the representations of its neighbors and itself, computed in the previous layer. This aggregation involves a linear transformation using the trainable weight matrices $W^{(0)},\ldots,W^{(L-1)}$ and the activation $\text{ReLU}(x) = \max(x,0)$. Therefore, assuming $L = 2$ as in the model of Kipf and Welling~\cite{kipf2016-2}, the VGAE encoding step of Equation~\eqref{neweq1} reformulates as follows:
\begin{equation}
\mu = \tilde{A} \text{ReLU} (\tilde{A} X W^{(0)}_{\mu}) W^{(1)}_{\mu}  \text{and } \Sigma = \tilde{A} \text{ReLU} (\tilde{A} X W^{(0)}_{\Sigma}) W^{(1)}_{\Sigma}.
\label{gcn1}
\end{equation}
Let $d_h \in \mathbb{N}^*$ denote the hidden layer dimension. Then, $W^{(0)}_{\mu}$, $W^{(0)}_{\Sigma}$ are $f \times d_h$ matrices (or $n \times d_h$, if $X = I_n$), while $W^{(1)}_{\mu}$, $W^{(1)}_{\Sigma}$ are $d_h \times d$ matrices.
Finally, node embedding vectors are sampled using these learned parameters. Letting $Z$ be the $n \times d$ matrix stacking all vectors $z_i \in \mathbb{R}^d$, Kipf and Welling~\cite{kipf2016-2} consider a mean-field model for sampling: $q(Z \mid A,X) = \prod_{i=1}^n q(z_i\mid A,X)$ with $q(z_i \mid A,X) = \mathcal{N}(z_i\mid \mu_i, \Sigma_i)$, where $\mathcal{N}(\cdot)$ is the Gaussian distribution.

\subsubsection{VGAE Decoding} 
To evaluate the encoding quality, the VGAE aims to decode, i.e., reconstruct the graph from embedding vectors. Kipf and Welling~\cite{kipf2016-2} rely on inner products to reconstruct edges, yielding the $n \times n$ reconstructed adjacency 
matrix $\hat{A}$ with:
\begin{equation}
\hat{A}_{ij} = p(A_{ij} = 1 \mid z_i, z_j) = s(z_i^\top z_j),
\end{equation} where $s(x) = 1/(1 + e^{-x})$ is the sigmoid function. This leads to the generative model $p(A \mid Z, X) = \prod_{i=1}^n  \prod_{j=1}^n  p(A_{ij} \mid z_i, z_j).$

During training, weight matrices of GCN encoders are tuned via gradient ascent to maximize the evidence lower bound (ELBO)~\cite{kingma2013vae}, a variational lower bound of the model's likelihood defined~as~:
\begin{equation}
\mathcal{L} = \mathbb{E}_{q(Z \mid A,X)} [\log
p(A \mid Z,X)] - \mathcal{D}_{KL}[q(Z \mid A,X)||p(Z)].
\end{equation}
The left term measures reconstruction quality and would be maximized when $\hat{A} = A$. Intuitively, the ability to accurately reconstruct a graph from the  embedding space suggests that this space effectively summarizes key information about the original structure~\cite{salha2022contributions}. The right term, in which $\mathcal{D}_{KL}[\cdot||\cdot]$ represents the Kullback-Leibler divergence~\cite{kullback1951information} and $p(Z)$ a unit Gaussian prior, acts as a regularization, controlling the magnitude of node embedding~vectors~\cite{kipf2016-2}.

\subsection{Weight Sharing in VGAE}
\label{s22}

\subsubsection{WS} This paper investigates weight sharing (WS) in VGAE, which refers to using the same weights across hidden layers of both encoders, i.e., all weights except those of the output layer. For instance, in the 2-layer GCNs of Equation~\eqref{gcn1}, WS consists~in~setting:
\begin{equation}
W^{(0)}_{\mu} = W^{(0)}_{\Sigma} = W^{(0)}.   
\label{ws}
\end{equation}

\subsubsection{Usage of WS in VGAE}
WS was introduced by default in the original TensorFlow implementation\footnote{GCNModelVAE class in: \href{https://github.com/tkipf/gae/blob/master/gae/model.py}{https://github.com/tkipf/gae}.} of VGAE~\cite{kipf2016-2}. The paper only briefly mentions this practice, noting that GCN encoders \textit{``share first-layer parameters''} \cite{kipf2016-2} in a single sentence, without further justification. More recently, WS in VGAE was also implemented in PyTorch Geometric\footnote{VariationalGCNEncoder class in: \href{https://github.com/pyg-team/pytorch_geometric/blob/master/examples/autoencoder.py}{https://github.com/pyg-team/pytorch\_geometric}.}, a popular PyTorch library for graph representation learning \cite{pytorchgeometric}. No additional rationale for the use of WS could be found in PyTorch Geometric's paper or code.

Most recent VGAE models have built on either one of these two implementations and, consequently, have also adopted WS, often without explicit acknowledgment. 
This has sometimes caused discrepancies between the descriptions of encoders in papers, which omit WS, and their actual implementations in code. For example, all VGAE variants discussed in Section~\ref{s3} have inherited WS. In practice, WS has quietly become the standard in the field, but no in-depth investigation of its relevance has been conducted to date.

\subsection{To Share or Not to Share?}
\label{s23}

We argue that the question of whether WS should be adopted is not straightforward, as it presents both advantages and drawbacks. 

\subsubsection{Advantages}
Several studies have examined WS in other neural network architectures beyond VGAE, highlighting its benefits. WS reduces the number of trainable parameters, thereby lowering memory and computational requirements, and facilitating deployment in resource-constrained environments~\cite{ws0,ws1,ws2,ws3,hamilton2020graph}.

We argue that this parameter efficiency aspect is particularly relevant for the GCN-based VGAE of Section~\ref{s21}, especially for featureless graphs as $W^{(0)}_{\mu}$, $W^{(0)}_{\Sigma}$ are $n \times d_h$ matrices, whereas $W^{(1)}_{\mu}$, $W^{(1)}_{\Sigma}$ are $d_h \times d$~matrices. 
Let us explore an illustrative example. Consider a graph with $n~=~5000$ nodes, with $d_h = 32$ and $d = 16$. In this case, $W_0$ matrices would contain $5000 \times 32 = 160000$ weights, whereas $W_1$ matrices would only include $32 \times 16 = 512$ weights. WS across $W^{(0)}_{\mu}$~and~$W^{(0)}_{\Sigma}$ would, therefore, allow sharing the vast majority of weights, significantly improving parameter~efficiency.

Additionally, WS simplifies VGAE implementations, streamlining model design and enabling easier reuse of pre-trained parameters. WS is also valued for its regularization effect, helping to reduce overfitting and promoting smoother optimization by imposing shared constraints on the parameter space. This enhances generalization by acting as an implicit inductive bias, encouraging consistent representations from both encoders~\cite{ws0,ws1,hamilton2020graph}.

\subsubsection{Drawbacks}
Despite these advantages, WS in VGAE also has several drawbacks. One key issue is the risk of over-constraining the model and reducing flexibility~\cite{ws0,ws1}. Using WS across encoders limits the ability to capture distinct patterns for means and variances, potentially leading to suboptimal representations~\cite{hamilton2020graph}.

Overall, WS also leads to a loss of expressiveness in models.
Consider incorporating the equality from Equation~\eqref{ws} into the GCN encoding step of Equation~\eqref{gcn1}. This step simplifies as follows:
\begin{equation}
\mu = B W^{(1)}_{\mu} \text{and } \Sigma = B W^{(1)}_{\Sigma},
\end{equation}
with:
\begin{equation}
B = \tilde{A} \text{ReLU} (\tilde{A} X W^{(0)}).
\end{equation}
In other words, using WS, $\mu$~and $\Sigma$ would become mere linear transformations of the same $n \times d_h$ matrix $B$, limiting the model's expressiveness compared to non-WS~settings.

Therefore, WS may cause performance degradation in tasks like link prediction and community detection. Indeed, by sharing most weights and limiting expressiveness and flexibility, a VGAE may struggle to capture complex data patterns, reducing its practical effectiveness.
To conclude this section, it remains unclear whether the benefits of WS outweigh its drawbacks, leaving the question of its adoption warranting further experimental~investigation.

\section{Experimental Analysis}
\label{s3}

This section presents our experimental analysis. Our source code is publicly available on GitHub, along with download information for all public datasets: \href{https://github.com/kiboryoku/ws_vgae}{https://github.com/kiboryoku/ws\_vgae}.

\subsection{Experimental Setting}

\subsubsection{Tasks}

We consider two evaluation tasks. The first one, link prediction, involves predicting the location of missing edges in graphs~\cite{kipf2016-2,liben2007link}. We train all VGAE models described thereafter, with and without WS, on incomplete graphs with $15\%$ of edges randomly masked. Then, we construct validation and test sets using these masked edges ($5\%$ and $10\%$, respectively) and an equal number of randomly selected unconnected node pairs. We evaluate each VGAE's performance in classifying edges versus non-edges using the decoded value $\hat{A}_{ij}$, as measured by Area Under the ROC Curve (AUC) and Average Precision (AP) scores~\cite{kipf2016-2,fawcett2006introduction}, averaged over 100 runs with different edge splits to reflect randomness.

The second task is community detection~\cite{salhagalvan2022modularity,fortunato2010community}, which involves clustering nodes into similar subgroups. As detailed below, some graphs in our experiments have node-level community labels. In this case, we train models on the entire graph, then run a $k$-means~\cite{ikotun2023k} in the embedding space to cluster the $z_i$ vectors. Then, we compare our $k$-means clusters to the true node-level community labels, using the Adjusted Mutual Information (AMI) and Adjusted Rand Index (ARI) \cite{vinh2010information,hubert1985comparing} as evaluation scores, averaged over~100~runs.

\subsubsection{Datasets} 
We consider 12 graphs--or 16, by treating feature and featureless variants of graphs having node features as separate datasets. Table~\ref{table1} reports the number of nodes ($n$), edges ($m$), communities ($k$, if available), and feature dimensions ($f$, if available) for~each~graph. They range from 877 nodes to 875713 nodes and vary in nature. 10 out of 16 graphs have community labels. 

We first examine the Cora, Citeseer, and Pubmed citation networks used by Kipf and Welling~\cite{kipf2016-2}, where nodes represent articles citing each other, described by bag-of-words features, and grouped into communities based on topics.
We add seven graphs from a previous VGAE study~\cite{salha2020simple}, with sources provided therein. They comprise two other citation networks, Arxiv-HepTh and a larger version of Cora referred to as Cora Large; the Hamsterster social network; and four web graphs of hyperlink edges connecting web pages: WebKD (with and without node features), Blogs (political blog pages falling into left-leaning or right-leaning communities), a medium-size Google web graph denoted Google, and a larger one denoted Google Large. We also include two graphs from other VGAE studies: SBM~\cite{salhagalvan2022modularity}, a synthetic graph with communities, generated by a stochastic block model~\cite{abbe2017community}; and Artists~\cite{salha2021cold}, a co-listening graph extracted from the recommender system of a global music streaming service. Nodes are music artists, connected based on co-listening data, and grouped into communities by country.

\subsubsection{Models}
\label{models}
For these tasks and graphs, we compare the performance of 10 different VGAE models with and without WS. First, we consider the original VGAE~\cite{kipf2016-2} with 2-layer GCN encoders and an inner product decoder, referred to simply as VGAE in the remainder of this section. We also test four variants: ARGVA~\cite{pan2018arga}, adding adversarial regularization; G-VGAE~\cite{salha2019-2}, a gravity-inspired VGAE with physics-based decoding; MA-VGAE~\cite{salhagalvan2022modularity,salha2022new}, a modularity-aware VGAE for improved community detection; and Graphite-VGAE~\cite{grover2019graphite}, employing iterative decoding. 
Finally, we study deeper versions of these models, denoted Deep VGAE, Deep ARGVA, Deep G-VGAE, Deep MA-VGAE, and Deep Graphite-VGAE. These models adopt 3-layer GCN encoders, as opposed to 2-layer encoders in their standard counterparts. We apply WS to both~hidden~layers.

We train models for 300 iterations using the Adam optimizer~\cite{kingma2014adam}. For large graphs ($n >$ 20000), we employ the FastGAE technique for scalability~\cite{salha2021fastgae}, decoding degree-based sampled subgraphs of 5000 nodes during training. We set $d = 16$ and use hidden layer(s) with dimension $d_h = 32$. We select other hyperparameters via grid search on link prediction validation sets. Notably, we test learning rates in the range $\{0.01, 0.02, \dots, 0.1\}$. For brevity, we provide the optimal values for each model in our GitHub repository.


\begin{table*}[!ht]
  \centering
\caption{Link prediction and community detection on all graphs using VGAE and Deep VGAE, with and without WS. Link prediction scores (AUC, AP) are computed on test sets. Community detection scores (AMI, ARI) are reported only for graphs with communities. Scores are averaged over 100 runs and reported with standard deviations to reflect randomness. Scores should be compared pairwise (WS vs. No WS). Results show that all WS scores fall within one standard deviation of the No~WS~scores.} 
 \resizebox{\textwidth}{!}{
  \begin{tabular}{c|cccc|cccc|cccc|cccc}
    \toprule
         \multirow{3}{*}{\textbf{Model}} & \multicolumn{4}{c}{\textbf{Blogs}} & \multicolumn{4}{c}{\textbf{Cora}} & \multicolumn{4}{c}{\textbf{Cora, with features}} & \multicolumn{4}{c}{\textbf{Cora Large}} \\
          & \multicolumn{4}{c}{($n$ = 1 224,  $m$ = 19 025, $k$ = 2)} & \multicolumn{4}{c}{($n$ = 2 708, $m$ = 5 429, $k$ = 6)} & \multicolumn{4}{c}{($n$ = 2 708, $m$ = 5 429, $f$ = 1 433, $k$ = 6)} & \multicolumn{4}{c}{($n$ = 23 166, $m$ = 91 500, $k$ = 70)} \\
     \cmidrule{2-17}
       & \ \textbf{AUC (in \%)} & \ \textbf{AP (in \%)} & \ \textbf{AMI (in \%)} & \ \textbf{ARI (in \%)} & \ \textbf{AUC (in \%)} & \ \textbf{AP (in \%)} & \ \textbf{AMI (in \%)} & \ \textbf{ARI (in \%)} & \ \textbf{AUC (in \%)} & \ \textbf{AP (in \%)} & \ \textbf{AMI (in \%)} & \ \textbf{ARI (in \%)} & \ \textbf{AUC (in \%)} & \ \textbf{AP (in \%)} & \ \textbf{AMI (in \%)} & \ \textbf{ARI (in \%)}  \\
    \midrule
    \midrule
    \textbf{VGAE - WS}  & 91.56 $\pm$ 0.63 & 92.61 $\pm$ 0.59 & 73.93 $\pm$ 0.63 & 83.12 $\pm$ 0.61 & 84.86 $\pm$ 1.48 &  88.67 $\pm$ 1.11 & 35.11 $\pm$ 2.89 & 26.05 $\pm$ 3.52 & 91.08 $\pm$ 0.92 & 92.38 $\pm$ 0.84 & 44.42 $\pm$ 2.58 & 34.17 $\pm$ 4.71 & 95.08 $\pm$ 0.17 & 95.94 $\pm$ 0.15 & 42.39 $\pm$ 0.59 & 15.58 $\pm$ 0.64 \\ 
     \textbf{VGAE - No WS}  & 91.59 $\pm$ 0.51 & 92.65 $\pm$ 0.50 & 73.83 $\pm$ 0.81 & 82.94 $\pm$ 0.65 &  85.05 $\pm$ 1.29 & 88.75 $\pm$ 0.96 & 34.95 $\pm$ 2.43 & 26.32 $\pm$ 3.77 & 90.87 $\pm$ 1.16 & 92.30 $\pm$ 1.11 & 44.35 $\pm$ 2.74 & 33.50 $\pm$ 4.74 & 95.08 $\pm$ 0.18 & 95.96 $\pm$ 0.17 & 42.25 $\pm$ 0.58 & 15.49 $\pm$ 0.70 \\ 
     \midrule
      \textbf{Deep VGAE - WS}  & 91.79 $\pm$ 0.59 &  92.67 $\pm$ 0.55 & 72.79 $\pm$ 0.66 & 82.19 $\pm$ 0.53 &  85.33 $\pm$ 1.35 & 88.58 $\pm$ 1.12 & 34.68 $\pm$ 2.66 & 26.03 $\pm$ 3.86 & 89.75 $\pm$ 1.14 & 90.91 $\pm$ 1.07 & 43.95 $\pm$ 3.89 & 34.68 $\pm$ 4.21 & 94.87 $\pm$ 0.19 & 95.56 $\pm$ 0.20 & 39.63 $\pm$ 0.68 & 13.33 $\pm$ 0.70 \\ 
     \textbf{Deep VGAE - No WS} & 91.62 $\pm$ 0.57 & 92.47 $\pm$ 0.56 & 72.84 $\pm$ 0.87 & 82.21 $\pm$ 0.72 &  85.02 $\pm$ 1.16 & 88.42 $\pm$ 1.01 & 35.12 $\pm$ 3.04 & 26.17 $\pm$ 4.10 & 89.78 $\pm$ 1.12 & 90.99 $\pm$ 1.12 & 43.13 $\pm$ 2.51 & 33.57 $\pm$ 3.92 & 94.88 $\pm$ 0.19 & 95.59 $\pm$ 0.19 & 39.67 $\pm$ 0.77 & 13.33 $\pm$ 0.65 \\ 
    \midrule
    \midrule
         \multirow{3}{*}{\textbf{Model}} & \multicolumn{4}{c}{\textbf{Citeseer}} & \multicolumn{4}{c}{\textbf{Citeseer, with features}} & \multicolumn{4}{c}{\textbf{Pubmed}} & \multicolumn{4}{c}{\textbf{Pubmed, with features}} \\
         & \multicolumn{4}{c}{ ($n$ = 3 327, $m$ = 4 732, $k$ = 7)} & \multicolumn{4}{c}{($n$ = 3 327, $m$ = 4 732, $f$ = 3 703, $k$ = 7)} & \multicolumn{4}{c}{ ($n$ = 19 717, $m$ = 44 338, $k$ = 3)} & \multicolumn{4}{c}{($n$ = 19 717, $m$ = 44 338, $f$ = 500, $k$ = 3)} \\
     \cmidrule{2-17}
       & \ \textbf{AUC (in \%)} & \ \textbf{AP (in \%)} & \ \textbf{AMI (in \%)} & \ \textbf{ARI (in \%)} & \ \textbf{AUC (in \%)} & \ \textbf{AP (in \%)} & \ \textbf{AMI (in \%)} & \ \textbf{ARI (in \%)} & \ \textbf{AUC (in \%)} & \ \textbf{AP (in \%)} & \ \textbf{AMI (in \%)} & \ \textbf{ARI (in \%)} & \ \textbf{AUC (in \%)} & \ \textbf{AP (in \%)} & \ \textbf{AMI (in \%)} & \ \textbf{ARI (in \%)}  \\
    \midrule
    \midrule
    \textbf{VGAE - WS}  & 78.04 $\pm$ 1.71 & 83.51 $\pm$ 1.30 & 9.48 $\pm$ 1.76 & 5.34 $\pm$ 1.69 & 88.51 $\pm$ 1.34 & 89.22 $\pm$ 1.49 & 20.08 $\pm$ 3.09 & 13.83 $\pm$ 4.83 & 85.10 $\pm$ 0.38 & 88.32 $\pm$ 0.33 & 17.84 $\pm$ 3.27 & 9.66 $\pm$ 4.02 & 96.27 $\pm$ 0.19 & 96.53 $\pm$ 0.17 & 30.23 $\pm$ 2.89 & 29.74 $\pm$ 2.99 \\ 
     \textbf{VGAE - No WS}  & 78.14 $\pm$ 1.63 & 83.62 $\pm$ 1.34 & 9.37 $\pm$ 1.50 & 5.24 $\pm$ 1.41 &  88.57 $\pm$ 1.34 & 89.31 $\pm$ 1.53 & 20.30 $\pm$ 4.06 & 14.67 $\pm$ 5.03 & 84.89 $\pm$ 0.41 & 88.18 $\pm$ 0.32 & 17.88 $\pm$ 3.19 & 8.59 $\pm$ 3.39 & 96.29 $\pm$ 0.17 & 96.55 $\pm$ 0.16 & 29.67 $\pm$ 3.61 & 29.03 $\pm$ 3.40 \\ 
     \midrule
      \textbf{Deep VGAE - WS}  & 77.88 $\pm$ 1.44 & 82.95 $\pm$ 1.16 & 9.79 $\pm$ 1.73 & 4.98 $\pm$ 1.38 &  87.05 $\pm$ 1.52 & 87.86 $\pm$ 1.62 & 21.86 $\pm$ 2.96 & 17.29 $\pm$ 3.95 & 85.56 $\pm$ 0.37 & 88.54 $\pm$ 0.30 & 19.14 $\pm$ 3.40 & 13.16 $\pm$ 5.76 & 95.92 $\pm$ 0.24 & 96.27 $\pm$ 0.22 & 29.31 $\pm$ 3.09 & 27.67 $\pm$ 4.42 \\ 
     \textbf{Deep VGAE - No WS} & 78.27 $\pm$ 1.38 & 83.30 $\pm$ 1.11 & 9.43 $\pm$ 1.89 & 5.11 $\pm$ 1.69 & 87.22 $\pm$ 1.66 & 87.94 $\pm$ 1.69 & 19.79 $\pm$ 2.87 & 14.84 $\pm$ 3.98 & 85.47 $\pm$ 0.38 & 88.51 $\pm$ 0.31 & 18.52 $\pm$ 3.39 & 12.07 $\pm$ 3.96 & 95.90 $\pm$ 0.27 & 96.28 $\pm$ 0.25 & 28.76 $\pm$ 3.04 & 26.85 $\pm$ 4.85 \\ 
    \midrule
    \midrule
         \multirow{3}{*}{\textbf{Model}} & \multicolumn{4}{c}{\textbf{WebKD}} & \multicolumn{4}{c}{\textbf{WebKD, with features}} & \multicolumn{4}{c}{\textbf{Hamsterster}} & \multicolumn{4}{c}{\textbf{SBM}} \\
         & \multicolumn{4}{c}{($n$ = 877, $m$ = 1 608)} & \multicolumn{4}{c}{($n$ = 877, $m$ = 1 608, $f$ = 1 703)} & \multicolumn{4}{c}{($n$ = 1 858, $m$ = 12 534)} & \multicolumn{4}{c}{($n$ = 100 000, $m$ = 1 498 844, $k$ = 100)} \\
     \cmidrule{2-17}
       & \ \textbf{AUC (in \%)} & \ \textbf{AP (in \%)} & \ \textbf{AMI (in \%)} & \ \textbf{ARI (in \%)} & \ \textbf{AUC (in \%)} & \ \textbf{AP (in \%)} & \ \textbf{AMI (in \%)} & \ \textbf{ARI (in \%)} & \ \textbf{AUC (in \%)} & \ \textbf{AP (in \%)} & \ \textbf{AMI (in \%)} & \ \textbf{ARI (in \%)} & \ \textbf{AUC (in \%)} & \ \textbf{AP (in \%)} & \ \textbf{AMI (in \%)} & \ \textbf{ARI (in \%)}  \\
    \midrule
    \midrule
    \textbf{VGAE - WS}  & 76.45 $\pm$ 3.59 & 82.52 $\pm$ 2.63 & -- & -- &  83.73 $\pm$ 3.98 & 86.57 $\pm$ 2.88 & -- & -- & 92.83 $\pm$ 0.52 & 93.94 $\pm$ 0.39 & -- & -- & 81.21 $\pm$ 0.33 &  84.04 $\pm$ 0.65 & 29.41 $\pm$ 1.88 & 5.75 $\pm$ 0.64 \\ 
     \textbf{VGAE - No WS}  & 76.08 $\pm$ 3.42 & 82.29 $\pm$ 2.43 & -- & -- &  84.15 $\pm$ 4.09 & 87.18 $\pm$ 2.71 & -- & -- & 92.87 $\pm$ 0.47 & 94.01 $\pm$ 0.48 & -- & -- & 81.17 $\pm$ 0.37 & 83.89 $\pm$ 0.76 & 27.81 $\pm$ 2.52 & 5.42 $\pm$ 0.95 \\ 
     \midrule
      \textbf{Deep VGAE - WS}  & 75.57 $\pm$ 3.13 & 81.81 $\pm$ 2.34 & -- & -- &  82.68 $\pm$ 4.66 & 86.25 $\pm$ 3.17 & -- & -- & 92.24 $\pm$ 0.72 & 93.25 $\pm$ 0.66 & -- & -- & 77.66 $\pm$ 0.81 & 76.69 $\pm$ 0.89 & 13.72 $\pm$ 4.59 & 1.72 $\pm$ 0.82 \\ 
     \textbf{Deep VGAE - No WS} & 77.09 $\pm$ 3.49 & 83.09 $\pm$ 2.37 & -- & -- &  83.54 $\pm$ 4.59 & 86.83 $\pm$ 3.15 & -- & -- & 92.29 $\pm$ 0.56 & 93.35 $\pm$ 0.57 & -- & -- & 77.73 $\pm$ 0.90 & 77.04 $\pm$ 1.07 & 15.43 $\pm$ 2.85 & 2.13 $\pm$ 0.79 \\ 
    \midrule
    \midrule
         \multirow{3}{*}{\textbf{Model}} & \multicolumn{4}{c}{\textbf{Arxiv-HepTh}} & \multicolumn{4}{c}{\textbf{Google}} & \multicolumn{4}{c}{\textbf{Google Large}}  & \multicolumn{4}{c}{\textbf{Artists}} \\
         & \multicolumn{4}{c}{($n$ = 27 770, $m$ = 352 807)} & \multicolumn{4}{c}{($n$ = 15 763, $m$ = 171 206)} & \multicolumn{4}{c}{($n$ = 875 713, $m$ = 5 105 039)} & \multicolumn{4}{c}{($n$ = 24 270, $m$ = 419 607, $k$ = 20)} \\
     \cmidrule{2-17}
       & \ \textbf{AUC (in \%)} & \ \textbf{AP (in \%)} & \ \textbf{AMI (in \%)} & \ \textbf{ARI (in \%)} & \ \textbf{AUC (in \%)} & \ \textbf{AP (in \%)} & \ \textbf{AMI (in \%)} & \ \textbf{ARI (in \%)} & \ \textbf{AUC (in \%)} & \ \textbf{AP (in \%)} & \ \textbf{AMI (in \%)} & \ \textbf{ARI (in \%)} & \ \textbf{AUC (in \%)} & \ \textbf{AP (in \%)} & \ \textbf{AMI (in \%)} & \ \textbf{ARI (in \%)}  \\
    \midrule
    \midrule
    \textbf{VGAE - WS}  & 96.48 $\pm$ 0.12 & 97.02 $\pm$ 0.10 & -- & -- &  94.38 $\pm$ 0.37 & 95.77 $\pm$ 0.32 & -- & -- & 95.15 $\pm$ 0.13 & 96.04 $\pm$ 0.14 & -- & -- & 98.07 $\pm$ 0.07 & 97.70 $\pm$ 0.08 & 33.29 $\pm$ 0.81 & 8.30 $\pm$ 0.45 \\ 
     \textbf{VGAE - No WS}  & 96.52 $\pm$ 0.12 & 97.04 $\pm$ 0.11 & -- & -- &  94.30 $\pm$ 0.33 & 95.69 $\pm$ 0.30 & -- & -- & 95.16 $\pm$ 0.10 & 96.05 $\pm$ 0.12 & -- & -- & 98.05 $\pm$ 0.07 & 97.68 $\pm$ 0.09 & 33.55 $\pm$ 0.68 & 8.57 $\pm$ 0.65 \\ 
     \midrule
      \textbf{Deep VGAE - WS}  & 92.77 $\pm$ 3.37 & 93.59 $\pm$ 3.24 & -- & -- &  93.63 $\pm$ 0.67 & 95.22 $\pm$ 0.50 & -- & -- & 95.06 $\pm$ 0.09 & 95.83 $\pm$ 0.08 & -- & -- & 97.44 $\pm$ 0.16 & 96.83 $\pm$ 0.22 & 32.58 $\pm$ 1.03 & 7.97 $\pm$ 0.74 \\ 
     \textbf{Deep VGAE - No WS} & 93.04 $\pm$ 2.99 & 93.79 $\pm$ 3.04 & -- & -- &  93.66 $\pm$ 0.54 & 95.26 $\pm$ 0.41 & -- & -- & 95.00 $\pm$ 0.09 & 95.78 $\pm$ 0.09 & -- & -- & 97.42 $\pm$ 0.10 & 96.74 $\pm$ 0.10 & 31.83 $\pm$ 0.81 & 7.73 $\pm$ 0.62 \\ 
    \bottomrule
  \end{tabular}
  }
  \label{table1}
\end{table*} 

\begin{table}[!ht]
  \centering
\caption{We ran experiments similar to those in Table~\ref{table1} for eight additional VGAE variants. The columns ``WS = No WS?'' indicate ``Yes'' if, as in Table~\ref{table1}, all scores obtained with WS fall within one standard deviation of those without WS.}
 \resizebox{\columnwidth}{!}{
  \begin{tabular}{c|c||c|c}
    \toprule
            \multirow{2}{*}{\textbf{Model}}& \multirow{2}{*}{\textbf{WS = No WS ?}} & \multirow{2}{*}{\textbf{Model}} & \multirow{2}{*}{\textbf{WS = No WS ?}} \\
            & & & \\
         \midrule
         \midrule
\textbf{ARGVA}~\cite{pan2018arga} & Yes & \textbf{G-VGAE}~\cite{salha2019-2} & Yes \\
\midrule
\textbf{Deep ARGVA}~\cite{pan2018arga} & Yes & \textbf{Deep G-VGAE}~\cite{salha2019-2} & Yes \\
\midrule
\textbf{Graphite-VGAE}~\cite{grover2019graphite} & Yes & \textbf{MA-VGAE}~\cite{salhagalvan2022modularity} & Yes \\
\midrule
\textbf{Deep Graphite-VGAE}~\cite{grover2019graphite} & Yes & \textbf{Deep MA-VGAE}~\cite{salhagalvan2022modularity} & Yes \\

\bottomrule
  \end{tabular} 
  }
  \label{table2}
\end{table}

\subsection{Results and Discussion}

\subsubsection{Results}

Table~\ref{table1} presents results for VGAE with and without WS. The findings are clear: across all graphs and both tasks, VGAE with WS consistently performs similarly to VGAE without WS, with all WS scores falling within one standard deviation of their non-WS counterparts.
For example, on Blogs, VGAE with WS achieves an AMI of 
$73.93\% \pm 0.63$, compared to $73.83\% \pm 0.81$ without WS.
Similarly, VGAE with WS records an AUC of $91.56\% \pm 0.63$, closely matching the $91.59\% \pm 0.51$ AUC without WS.
In none of the settings does enforcing WS lead to performance deterioration.
Table~\ref{table1} shows that these conclusions also hold for Deep VGAE, while Table~\ref{table2} succinctly confirms these findings for the other models: ARGVA, G-VGAE, MA-VGAE, Graphite-VGAE, and their Deep variants.

\subsubsection{Robustness Checks}
During our tests, we also observed that WS models remain competitive with non-WS models when: (1) increasing training iterations up to 500; (2) adding one or two hidden layers; (3) increasing edge masking up to $20\%$ for link prediction; (4) using $d = 32$, $64$, or $128$ and $d_h = 64$ or $128$; and (5) comparing models using statistical confidence bounds instead of checking if scores are within a standard deviation from each other.  
These results, replicable using our code, confirm the robustness of our~conclusions.

\subsubsection{Discussion} While WS has both positive and negative implications, as discussed in Section~\ref{s2}, our experiments consistently show that its benefits outweigh its drawbacks. WS appears as an effective method for simplifying VGAE models, with no observed performance loss in any experiment. These findings hold across both tasks, 10 model variants, and 16 graphs of diverse sizes (ranging from hundreds to hundreds of thousands of nodes), types (including web graphs, citation networks, social networks, and co-listening graphs), and characteristics (with or without node features or node communities).
Relaxing WS to learn different hidden weights for Gaussian means and variances does not improve results, suggesting that, at least for our tasks, such additional complexity is unnecessary and likely excessive. Moreover, we note that, during our experiments, WS models usually achieved an average training time that was approximately 10\% shorter than that of models without WS, underscoring the efficiency gains of this approach.
For all these reasons, we confirm that WS is a relevant practice for VGAE models and recommend its use in future work to balance simplicity, efficiency, and performance in VGAE implementations.

\subsubsection{Limitations and Future Work} While we considered a wide range of configurations, our study does not cover all possible tasks (e.g., graph generation from node embedding vectors~\cite{zhu2022survey}) or models (e.g., VGAE models replacing GCNs with attention-based graph neural network encoders~\cite{velivckovic2019graph}, or VGAE models involving other probabilistic distributions such as Dirichlet distributions~\cite{li2020dirichlet} or Gaussian mixtures~\cite{choong2018learning,choong2020optimizing}). 
We hope that publishing our source code on GitHub will support future research on WS in these contexts. In our future work, we ourselves plan to broaden the scope of our experiments, to explore the use of WS in VGAE for dynamic graphs, and to conduct theoretical studies on the impact~of~WS.

\section{Conclusion}
\label{s4}

This paper demonstrates, through detailed analysis and comprehensive experiments, that the advantages of WS in VGAE models consistently outweigh its limitations. WS proves to be an effective approach for optimizing and simplifying VGAE models without performance loss across diverse settings. For these reasons, we advocate for its use in future work. Additionally, in this paper, we identify promising future research directions and provide our source code to encourage further studies on WS.

\bibliographystyle{ACM-Reference-Format}
\balance
\bibliography{sample-base}

\end{document}